\documentclass[runningheads]{llncs}
\usepackage{graphicx}
\usepackage{amsmath,amssymb} % define this before the line numbering.
\usepackage{color}
\usepackage[width=122mm,left=12mm,paperwidth=146mm,height=193mm,top=12mm,paperheight=217mm]{geometry}
\usepackage{hyperref}
\usepackage{multirow}
\usepackage{url}
\usepackage{color}
\usepackage{subcaption}
\usepackage[multiple]{footmisc}

\newcommand{\sigmoid}{\mbox{sigmoid}}

\newcommand{\var}[1]{{\operatorname{#1}}}
\begin{document}
% \linenumbers
%\pagestyle{headings}
\mainmatter

\title{Weakly Supervised Medical Diagnosis and Localization from Multiple Resolutions} % Replace with your title

%\titlerunning{ECCV-18 submission ID \ECCV18SubNumber}

%\authorrunning{ECCV-18 submission ID \ECCV18SubNumber}

\author{Li Yao, Jordan Prosky, Eric Poblenz, Ben Covington, Kevin Lyman\\
\texttt{\{li,prosky,eric,ben,kevin\}@enlitic.com}
}
\institute{Enlitic Inc. \\
San Francisco, CA 94111, USA\\
}
\maketitle

\begin{abstract}
Diagnostic imaging often requires the simultaneous identification of a multitude of findings of varied size and appearance. Beyond global indication of said findings, the prediction and display of localization information improves trust in and understanding of results when augmenting clinical workflow. Medical training data rarely includes more than global image-level labels as segmentations are time-consuming and expensive to collect. We introduce an approach to managing these practical constraints by applying a novel architecture which learns at multiple resolutions while generating saliency maps with weak supervision. Further, we parameterize the Log-Sum-Exp pooling function with a learnable lower-bounded adaptation (LSE-LBA) to build in a sharpness prior and better handle localizing abnormalities of different sizes using only image-level labels. Applying this approach to interpreting chest x-rays, we set the state of the art on 9 abnormalities in the NIH's CXR14 dataset while generating saliency maps with the highest resolution to date. %The code to completely reproduce the experiments is made available.

\end{abstract}
\section{Introduction}
\subsection{The Challenge of Diagnosing Abnormalities at Different Scales} \label{sec_scale}
Radiologists in clinical practice are responsible for the correct interpretation of all items on an image. In the interpretation of a chest x-ray, for example, they are in search of dozens of visual patterns indicative of hundreds of potential clinical outcomes. These patterns are often highly varied in their appearance and most easily discerned at varying levels of analysis. Enlargement of the cardiac silhouette, for example, is determined to be present when the width of the heart is measured to be 50\% or greater than the width of the thoracic cage--a pattern far more easily detected when viewing the entirety of the image rather than a set localized region. Lung nodules, by contrast, are often subtle findings as small as a few millimeters in size and are frequently missed by practitioners even when viewed closely on a high resolution monitor. Many patterns, such as those of interstitial changes, are presented in the form of visual features at both the macro and micro level. Diffuse infiltrative opacification in the periphery of the lung is often more easily noted from a global view and suggests the presence of the pattern itself, but closer inspection of the anomalous region is often required to narrow the differential diagnosis and determine followup. 

These phenomena suggest utility in image analysis at multiple levels of resolution, gaining value from both a global and highly localized set of views. This hypothesis is conceptually supported by observing the clinical workflow of diagnostic radiologists, who heavily rely on the ability to zoom in and out of images throughout various stages of their interpretation. 
\subsection{The Challenge of Obtaining Medical Interpretation Beyond Global Predictions}\label{sec_interpretation} While there is a great deal of utility in the global detection of anomalous patterns, it is often desirable to additionally localize these findings. Localization can be used to draw immediate visual attention to findings of interest, augmenting diagnostic workflows by enabling radiologists to provide faster and more accurate reads. Further, the black box€ nature of deep learning poses significant challenges in the adoption of these solutions into clinical practice; saliency maps help build trust from clinicians in offering a form of transparency into the processes which led to a given prediction. This insight is also useful in model development where saliency maps can often help indicate underlying biases which otherwise would be difficult to trace--a global prediction indicating the presence of a pneumothorax may be correct, but inspection of a corresponding saliency map may reveal a serendipitous false positive triggered by the presence of a different but visually similar feature like a vertically oriented chest tube or skin fold. 

Medical training data is very challenging to label as it often relies either on the use of natural language processing to convert historic reports into global labels, or the use of networks of medical experts to prospectively read and annotate studies. Segmentation information to accompany the assigned labels is impossible to attain with the former approach as this information is not captured in reports, and adds a great deal of expense in the latter as manually drawing segmentations is typically a time intensive endeavor. These factors incentivize the development of an approach which enables localization only from the use of global labels.
\subsection{Weakly Supervised Multi-Instance Learning}
The aforementioned challenges in automated and semi-automated medical diagnosis motivate the design of machine learning models that are capable of not only making global predictions but also providing pathology-based saliency maps that are clinically interpretable and insightful, and being able to do so under the strict constraint that bounding boxes or ROIs are rarely available. 

Such a problem is commonly formulated under the framework of multi-instance learning (MIL) (\cite{dietterich1997solving}, \cite{amores2013multiple}) where an image may contain zero or more types of instances and hence is treated as a bag. One of the most commonly used multi-instance learning assumptions assigns a positive class label to a bag if it contains any number of such instances and a negative label otherwise. Indeed, MIL appears in recent studies of \cite{chestx8} and \cite{feifei} for chest X-rays and \cite{courtiol2018classification} for histopathology,  all of which perform classification and localization without or with very limited local annotations. Beyond the domain of medicine, MIL has made its frequent appearance in various contexts such as those in Section \ref{sec_wsl}.
\subsection{Main Contributions}
\begin{figure}[ht]
\centering
\includegraphics[scale=0.33]{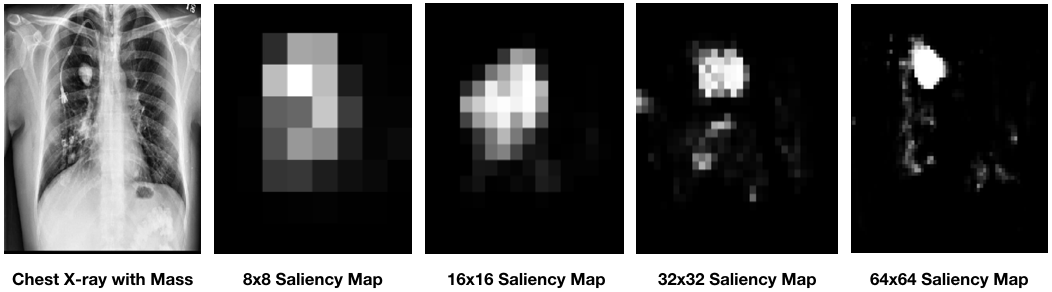}
\caption{Chest X-ray with saliency maps of increasing resolutions. With multi-resolution, lower-resolution maps provide weak localization cues that are refined in higher-resolution layers. The proposed model is therefore able to generate saliency maps of the highest resolution to date. Four models with an increasing target resolution were trained to produce the visualization.}
\label{overview}
\end{figure}

We argue, however, that existing models are not able to properly address the challenges posed from Section \ref{sec_scale} and Section \ref{sec_interpretation}. As reviewed in Section \ref{sec_multi_resolution}, previous work on localization with multi-resolution mainly focuses on the fully supervised setting where ground truth ROIs are directly utilized in training, exemplified by the dominant U-net (\cite{u_net}) family and its variants. This work instead studies multi-resolution with MIL where ROIs are not available. Unlike the previous related work of Section \ref{sec_wsl}, we parameterize the Log-Sum-Exp pooling function with a learnable lower-bounded adaptation (LSE-LBA) to build in a sharpness prior and better handle the challenge of localizing abnormalities of different sizes using only image-level labels. As a result, the proposed LSE-LBA pooling has the benefit of being self-adaptive and numerically more stable to compute when pooling in higher resolutions. This design allows our model to generate high-resolution, crisp probabilistic saliency maps in a principled fashion without relying on localization labels. Figure \ref{overview} illustrates the importance of high-resolution saliency maps for reliable clinical interpretation.
%\subsection{Main Contributions}
%Novel multi-resolution MIL cost to perform weakly-supervised localization, stable and flexible LSE-LBA pooling, SOTA performance?
\section{Related work}

\subsection{Multiple Resolution Methods in Computer Vision}\label{sec_multi_resolution}
Robust image understanding requires models that incorporate information from multiple scales and levels of abstraction. This is especially important when objects vary considerably in size or require context to recognize. Whereas traditional approaches used image pyramids \cite{pyramids}, most state-of-the-art models now rely on the layers of deep convolutional neural networks (CNNs) which can be seen as a non-linear counterpart of image pyramids \cite{hypercolumns}.

The feature maps in the last convolutional layer encode all of this information which can be used to produce image-level class predictions or pixel-level dense predictions in the form of saliency maps that visualize the image regions associated with each class \cite{fcn,class_activation_maps}. However, despite being informed by high-resolution features from the early layers, the last feature maps typically lack sufficient spatial resolution for precise localization and yield coarse saliency maps \cite{fcn}. This is due to downsampling operations in the CNN introduced to increase the receptive field of each feature while keeping the computational requirements in check.

To generate detailed saliency maps that precisely delineate both large and small objects, one must combine the output from multiple layers or resolutions which entails upsampling the low resolution feature maps to match the desired saliency map resolution. Various approaches to upsampling have been considered including bilinear interpolation \cite{class_activation_maps,refine_net} and learnable ``deconvolution'' filters \cite{fcn,u_net}. Badrinarayanan et al. \cite{seg_net} proposed to upsample feature maps decimated by max-pooling by using the indices of the max to position each low resolution value. The upsampling step is then followed by a series of convolutions to densify the maps. Our approach is similar in that it uses a deterministic nearest-neighbors upsampling step followed by a learnable non-linear refinement. This is both computationally efficient and compatible with a ResNet-type \cite{res_net} encoder which decimates by strided convolution.

How best to combine features learned at different resolutions remains an open question. One of the earliest approaches was to simply average the dense predictions from each upsampled layer \cite{fcn}. Another approach was to concatenate the upsampled features prior to classification \cite{hypercolumns}. However, most models now employ an incremental approach in which the features from convolutional layer $l$ are upsampled to match the resolution of layer $l-1$, combined with the feature maps from layer $l-1$, and then refined \cite{u_net,seg_net,tiramisu,refine_net}. The process is repeated until the resulting feature maps are sufficiently dense. This was introduced by the so-called U-net image segmentation model which emphasized the importance of the refinement in the upsampling path \cite{u_net,tiramisu}. Our model resembles the above but replaces the pixel-wise supervision with the image-wise multi-instance learning.
%J{\'{e}}gou et al.\ \cite{} were also motivated by the implicit deep supervision provided by DenseNet-type connections and proposed a similar upsampling path. Our model employs this incremental approach but with DenseNet-type \cite{huang2017densely} combination and refinement to encourage the learning of independent features that are shared both between and within layers.

\subsection{Weakly Supervised Learning}\label{sec_wsl}
Learning with weak supervision involves learning from incomplete, inexact, or inaccurate labels \cite{weaklearning}. This is a common problem in the domain of medical imaging, where it is costly to obtain pixel-level labels and it is desirable to be able to localize or segment abnormalities from image-level labels. There are many different approaches to solving the weakly supervised learning problem; a method that has garnered a lot of attention in recent years is MIL. In the MIL framework, training data consists of labeled bags (i.e. medical images), where a bag is composed of multiple instances (i.e. image patches). A standard assumption in MIL is that a bag should be labeled as positive if at least one of its instances is positive.

CNNs are a natural fit for the MIL framework because it has been shown that they effectively learn object detectors even when trained as a classifier using only global labels \cite{detectors_emerge}. Pathak \textit{et al.} \cite{fcnmil} used a fully-convolutional VGG-16 network to learn pixel-level segmentation from image-level labels by predicting an output map for all pixels and using max  pooling to reduce to a global label. Pinheiro \textit{et al.} \cite{im2pix} followed a similar approach, but utilizes the Overfeat \cite{overfeat} model to generate feature maps for MIL and uses a smoothed version of max pooling called Log-Sum-Exp. A key difference between many MIL approaches is the choice of pooling function, which is often a smooth approximation of the max function \cite{babenko}. In the medical imaging domain, Kraus \textit{et al.} \cite{microscopy} applied MIL to segment microscopy images with the use of an Adaptive Noisy-AND pooling function on CNN feature maps. Zhu \textit{et al.} \cite{mammomil} used a pooling function that involved ranking instances with the goal of performing end-to-end mass classification for whole mammograms, and Li \textit{et al.} \cite{feifei} used Noisy-OR pooling to identify and localize thoracic diseases. Recently, Ilse \textit{et al.} \cite{attentionmil} used a two-layered neural network to perform attention-based MIL pooling, providing adaptivity and flexibility.

MIL is just one approach to learn localization information from neural networks designed for classification. Zhou \textit{et al.} \cite{class_activation_maps} used a technique that generates an image which maximizes class score, as well as class saliency maps for the task of weakly supervised segmentation using classification CNNs. Later, Selvaraju \textit{et al.} \cite{gradcam} introduced gradient-weighted Class Activation Mapping (CAM) which uses gradients flowing to the final convolutional layer to produce a localization map. A key difference between these approaches and ours is that we explicitly train our model to localize, whereas CAM methods are used ad-hoc to try to force a trained classification network to output weak localization cues \cite{seedexpand}. 
% Moreover, Kolesnikov \textit{et al.} \cite{seedexpand} introduced loss functions with the goal of training a weakly supervised segmentation model.

% A common theme missing from previous work on MIL is a discussion of the effect of bag-size on performance. That is, there is sparsely sufficient rigor placed on determining the optimal feature map resolution on which to perform MIL. Many people pick low-resolution feature maps, but they can lose useful localization information by doing so, especially when the object of interest is very small. In this work, we examine pooling from different-sized feature maps and show that combining information across resolutions can improve classification and localization results. 

\subsection{Medical Deep Learning}\label{sec_medical_deep_learning}
Deep learning methods have recently become popular for the analysis of medical images. Applications include faster and more accurate diagnosis, more efficient triage systems, image quality enhancement, and more. In this work, we focus on the problem of classifying and localizing abnormalities in chest x-rays.

One of the first neural networks applied to biomedical image segmentation was U-net \cite{u_net}, which has been iterated upon to produce 3D variants such as V-net \cite{vnet}. Training such networks, however, usually requires pixel-level segmentation masks which are difficult and expensive to acquire. In order to circumvent the dependence on pixel-level masks, many researchers have investigated weakly-supervised learning to identify the location of ROIs in both natural and medical images. Jia \textit{et al.} \cite{deep_weak_supervision} used a fully convolutional network with MIL to segment cancer regions in histopathology images. Wang \textit{et al.} \cite{chestx8} used a similar approach to ours in order to detect and spatially locate thoracic diseases. The authors, however, only consider feature maps at one resolution, employ a non-adaptive pooling function, and rely on pre-trained networks for initialization. Li \textit{et al.} \cite{feifei} also classified and localized thoracic diseases, but akin to \cite{chestx8}, the authors relied on network outputs at one fixed resolution. Rajpukar \textit{et al.} \cite{chexnet} also attempted to localize diseases in chest x-rays, but they utilized class activation maps.

Other researchers have also attempted to improve the classification of thoracic diseases. Yao \textit{et al.} \cite{LiLearning} leveraged interdependencies among 14 pathologies in chest x-rays in order to more accurately classify them, while Guan \textit{et al.} \cite{agcnn} used attention guided CNNs that generated masks to crop ROIs which were then classified. Kumar \textit{et al.} \cite{cascade} cascaded multiple predictions using binary relevance to improve performance on the multi-label prediction task. Wang \textit{et al.} \cite{tienet} proposed utilizing additional radiology reports to improve image classification.

\begin{figure}[ht]
\centering
\includegraphics[scale=0.055]{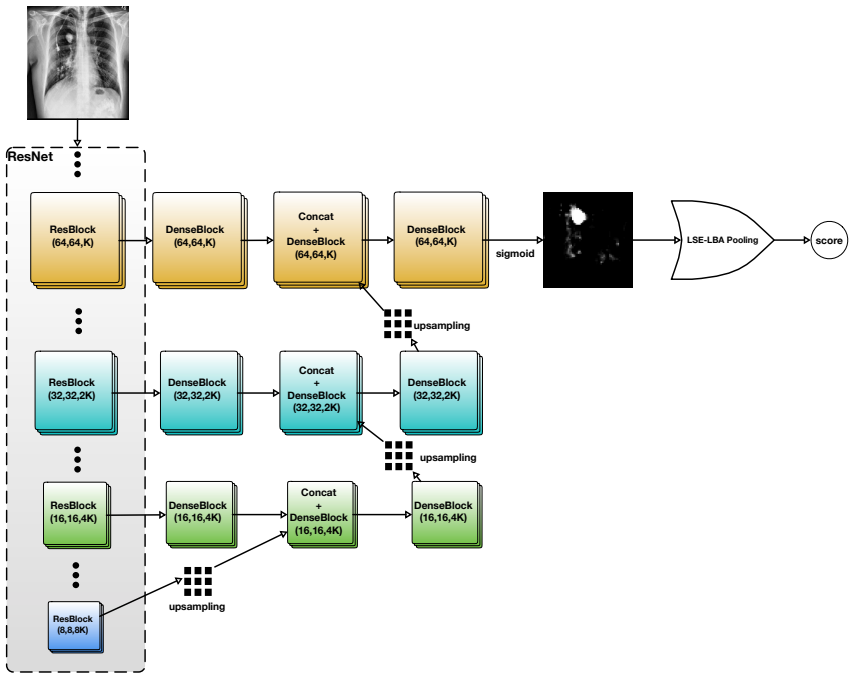}
\caption{The overall architecture of the proposed model from an input X-ray image to the predicted abnormality score. To reduce the resolution, a standard ResNet is firstly applied on the input image (indicated in the dotted rectangle box, Section \ref{sec_reduce}). To preserve the resolution, a standard DenseNet is applied per resolution (colored horizontal flow, Section \ref{sec_preserve}). Upsampling and channel-wise concatenation fuse information from multiple resolutions (Section \ref{sec_fuse}). $\var{\mbox{LSE}-{\mbox{LBA}}}$ pooling aggregates instance scores to the global probability (Section \ref{sec_mil}).}
\label{fig_model}
\end{figure}

\section{Proposed Model}
The following notations are used throughout the paper. Denote $\mathbf{x} \in \mathcal{R}^{w \times h \times c}$ as an input image with width $w$, height $h$ and channel $c$, $\mathbf{y}$ as a binary vector of dimensionality $K$, the total number of classes. The superscript indicates a specific dimension. For a specific class $k$, $\mathbf{y}^k=0$ indicates its absence and $\mathbf{y}^k=1$ its presence. The subscript indexes a particular example, for instance, $\{\mathbf{x}_i, \mathbf{y}_i\}$ is the $i$-th example. A feature map is $F \in \mathcal{R}^{w \times h \times c}$ while a saliency map is $S \in \mathcal{Q}^{w \times h \times K}$ with $\mathcal{Q} \in [0, 1]$. We also define two depth factors $l$ and $m$ to accompany feature and saliency maps. For instance, $F^l$ is the feature map as the result of a set of nonlinear transformation that changes the spatial resolution of $F^{l-1}$ . On the other hand, $F_{m-1}$ and $F_{m}$ are consecutive feature maps that preserve the resolution during the nonlinear transformation.

The proposed model is outlined in Figure \ref{fig_model}. For simplicity, the figure illustrates a design that produces a saliency map with a resolution of $64 \times 64$, $S^{0} \in \mathcal{Q}^{64 \times 64 \times 1}$, compared with $32 \times 32$, the highest resolution to date by \cite{chestx8}. The following sections describe a model parameterized for general multi-resolution multi-label classification.

\subsection{Reducing Resolutions with Residual Connections}\label{sec_reduce}
ResNets, first proposed in \cite{res_net} have been successfully utilized in various medical applications (Section \ref{sec_medical_deep_learning}). Compared with their predecessors, ResNets have the distinct advantage of being easy to optimize with gradient descent, largely due to their incorporation of residual connections. In particular, a ResNet contains several sub-modules, each of which is parameterized as $F^{l+1} = \sigma (g(F^{l}) + f(F^{l}))$, where $F^{l+1}$ is typically half the resolution of $F^{l}$ but has twice the number of channels, and $\sigma$ is an element-wise nonlinearity. The functions $g$ and $f$ are composed of a series of $1\times 1$ and $3 \times 3$ convolutions. The reduction in spatial resolution is achieved by using convolutions with a stride size 2. A key design in ResNets is choosing a simple $f$ and complex $g$ such that $f$ is as close as possible to a simple identity transformation, leaving the heavy-lifting non-linear transformations to $g$ to learn the residual. Optionally, \cite{res_net} inserts into ResNets building blocks preserving spatial resolutions with $F_{m+1} = \sigma (g(F_{m}) + F_{m})$ in which case $f$ is chosen to be the identity function. We adopt this convention.

\subsection{Refining Resolutions with Dense Connections}\label{sec_preserve}
ResNet is suitable as a standard off-the-shelf model for classification tasks. It is however susceptible to over-parameterization, as pointed out in \cite{wu2016wider} where the effective depth of a ResNet may be much more shallow, an indication that a large proportion of its parameters are underutilized. Such an issue becomes critical when residual connections are used repeatedly on the horizontal data flow in Figure \ref{fig_model} without changing the spatial resolution. In the scenario where  
\begin{align}\label{equ_resnet}
F^{l+1} = \sigma (g(F^{l}) + F^{l})
\end{align}
is applied repeatedly, a model could simply learn to ignore the capacity in $g$, especially when $\sigma=\mbox{relu}$. This effectively defeats the purpose of inner-resolution propagation where a model is encouraged to specialize in making predictions under a selected resolution $l$. 

The key to solve the above issue is to enforce explicitly the non-identity transformation on $F^{l}$, which suggests removing the residual connections. The resulting model would however lose the attraction of being easy to optimize. Such a trade-off naturally leads to the adoption of DenseNets from \cite{huang2017densely} where the resolution-preserving transformation is formulated as 
\begin{align}\label{equ_densenet}
F_{m+1} = \sigma (f(F_1 \oplus F_2 \oplus \dots \oplus F_m)) ,
\end{align}
where $\oplus$ denotes the channel-wise concatenation of feature maps and $f$ denotes a series of resolution-preserving nonlinear transformation. 

Compared with Equ. (\ref{equ_resnet}), Equ. (\ref{equ_densenet}) enforces the nonlinear transformation $f$ on all previous feature maps without the possibility of skipping using identity mapping while still maintaining the desirable property of being easy to optimize due to the direct connections with all previous feature maps. Such a design effectively encourages the participation of all previous feature maps in propagation.
\subsection{Combining Resolutions with Upsampling and Multi-Scale Dense Connections}\label{sec_fuse}
Fine-scale features, computed at high resolutions, capture detailed appearance information while coarse-scale features, computed from lower resolution representations of the data, capture semantic information and context. In deep neural networks, fine-scale features are learned in the earliest layers and coarse-scale features in the subsequent layers, where the spatial resolution of the data has been reduced by repeated downsampling operations. Thus, the model learns to construct a feature hierarchy in a fine-to-coarse manner.

While the coarse-scale features at the top of typical classification neural networks are suitable for image-level classification, spatial information required to precisely localize objects is likely to be lost. If the model is expected to predict not only what abnormalities are present in the image but where they are, then the spatial information must be reintegrated. Indeed, answering both of these questions is critical for most medical applications.

The proposed model (figure \ref{fig_model}) does this incrementally, in a coarse-to-fine manner, by repeatedly performing the following operation:
\begin{align}
F_{m}^{l} = f(\mathcal{U}(F_{n}^{l+1}) \oplus F_{m-1}^l) ,
\end{align}
where $F_{m}^{l}$ denotes the $m$-th resolution-preserving feature map at resolution level $l$, $F_{n}^{l+1}$ the $n$-th feature map from the lower resolution level $l+1$, $F_{m-1}^l$ the previous feature map at resolution level $l$ and $\oplus$ the channel-wise concatenation. The upsampling operation, $\mathcal{U}$, could be implemented in various ways including bilinear interpolation, nearest-neighbors interpolation, or even learnable transposed convolutions \cite{fcn}. Observing no significant performance difference among the choices, we opted to use the simplest approach which is nearest-neighbors. We argue that anything more expressive in the upsampling stage is not useful in our context because of the non-linearities we apply afterwards.

\subsection{The Choice of Pooling Function for Multi-instance Learning}\label{sec_mil}

Before introducing the overall cost function for our model, we first discuss pooling strategies to obtain an image-level label from each saliency map. Let $S \in \mathcal{Q}^{w \times h \times 1}$ be a 2D saliency map for a particular class $k$ to be pooled, and let $S_{i, j}$ be the $(\mbox{i}, \mbox{j})$-th element of $\mbox{S}$.

As discussed in section 2, there are many types of pooling functions which can be used in the MIL setting. Naive choices like max and average pooling tend to underestimate and overestimate the sizes of objects, respectively (\cite{seedexpand}), and it is therefore desirable to use a pooling function that provides more flexibility. One possible choice used by \cite{feifei} is the Noisy-OR (NOR) function, which is given by: $\mbox{NOR}(S) = 1 - \prod_{j} (1 - \mbox{p}_j)$, where j indexes grid positions in a 2D saliency map. A practical issue with Noisy-OR is numerical underflow which is a result of multiplying many small numbers together. Another choice is the generalized-mean (GM) function, which is given by: $\mbox{GM}(S) = ( \frac{1}{\mbox{wh}} \sum_{i=1}^{\mbox{w}} \sum_{j=1}^{\mbox{h}} (S_{i,j})^r )^{\frac{1}{r}}$. The hyper-parameter r controls the degree to which the GM function behaves more similar to max or average pooling. Indeed, as $\mbox{r} \to \infty$, GM essentially conducts max pooling, and it acts like average pooling as $\mbox{r} \to 0$. It is possible to convert the GM function to be adaptable, as \cite{adaptpool} did, where r becomes a learned-parameter during training. The GM function, though flexible, is also likely to suffer from underflow issues during computation as saliency map resolutions increase and many small numbers are multiplied. A function which is resilient to numerical issues and appears very similar to GM is Log-Sum-Exp, given by: $\mbox{LSE(S)} = \frac{1}{r} \log \{ \frac{1}{\mbox{wh}} \sum_{i=1}^w \sum_{j=1}^h \text{exp}[r\mbox{S}_{i,j}] \}$. 

\subsection{Log-Sum-Exp Pooling with Lower-bounded Adaptation ($\var{\mbox{LSE}-{\mbox{LBA}}}$)}\label{sec_lselba}
We introduce a variant of LSE pooling that takes $\mbox{S}$ and produces a final score $\mbox{p}$ as follows:
\begin{align}\label{lse}
\mbox{p} = \var{\mbox{LSE}-{\mbox{LBA}}}(\mbox{S}) = \frac{1}{r_{0} + \exp(\beta)}\log\{ \frac{1}{\mbox{wh}} \sum_{i=1}^{\mbox{w}} \sum_{j=1}^{\mbox{h}} \exp{[(r_{0} + \exp(\beta)) \mbox{S}_{i,j}}]\}.
\end{align}

% \subsubsection{$\var{\mbox{LSE}-{\mbox{LBA}}}$ is numerically more stable.} In addition to maintaining the benefits of using a pooling function which balances average- and max-pooling, the proposed pooling function is robust to the issue of numerical underflow when $S_{i,j}$ is very close to zero, compared with $\mbox{NOR}$ and $\mbox{GM}$ pooling, due to the removal of exponential that directly acts on $S_{i,j}$.

\noindent \textbf{$\var{\mbox{LSE}-{\mbox{LBA}}}$ is numerically more stable.} In addition to maintaining the benefits of using a pooling function which balances average and max pooling, the proposed pooling function is robust to the issue of numerical underflow when $S_{i,j}$ is very close to zero, compared with $\mbox{NOR}$ and $\mbox{GM}$ pooling, due to the removal of the exponential that directly acts on $S_{i,j}$.
\\
% \subsubsection{$\var{\mbox{LSE}-{\mbox{LBA}}}$ preserves probabilities.} A necessary point to address is that the range of this function is $\mathbb{R}$. If we use the output of the pooling function without being careful, the $\mbox{p}'s$ could not be interpreted as probabilities and an entropy-based loss would not make sense. Luckily, by bounding the values in $S$ to be in the range $[0, 1]$, the resulting score will also be in the same interval. Since the $\var{\mbox{LSE}-{\mbox{LBA}}}$ function is monotonically increasing in $S_{i,j}$, it attains its maximum value when all $S_{i,j} = 1$, and its minimum value when all $S_{i,j} = 0$. When S is a map of all 0's, $\var{\mbox{LSE}-{\mbox{LBA}}}(S) = \frac{1}{r}\log(1) = 0$, and when S is a map of all 1's, $\var{\mbox{LSE}-{\mbox{LBA}}}(S) = \frac{1}{r} \log(\exp{(r)}) = 1$. We use the sigmoid activation function on each $S_{i, j}$ to maintain this property.
\\
\noindent \textbf{$\var{\mbox{LSE}-{\mbox{LBA}}}$ preserves probabilities.} A necessary point to address is that the range of this function is $\mathbb{R}$. If we use the output of the pooling function without being careful, the $\mbox{p}'s$ could not be interpreted as probabilities and an entropy-based loss would not make sense. Luckily, by bounding the values in $S$ to be in the range $[0, 1]$, the resulting score will also be in the same interval. Since the $\var{\mbox{LSE}-{\mbox{LBA}}}$ function is monotonically increasing in $S_{i,j}$, it attains its maximum value when all $S_{i,j} = 1$, and its minimum value when all $S_{i,j} = 0$. When S is a map of all 0's, $\var{\mbox{LSE}-{\mbox{LBA}}}(S) = \frac{1}{r}\log(1) = 0$, and when S is a map of all 1's, $\var{\mbox{LSE}-{\mbox{LBA}}}(S) = \frac{1}{r} \log(\exp{(r)}) = 1$. We use the sigmoid activation function on each $S_{i, j}$ to maintain this property.
\\
%\subsubsection{Lower-bounding the adaptation of $r$.}
%The adaptive nature of $\var{\mbox{LSE}-{\mbox{LBA}}}$ is inspired by the design of adaptive \mbox{GM} pooling proposed in \cite{adaptpool}. In addition to being numerically more stable in computation, our proposed pooling function reparameterizes $r$ in $\mbox{LSE}$ pooling with $r = r_0 + \exp(\beta)$ where $r_0$ is a positive constant and $\beta$ a learnable parameter. It is easy to see that $r$ is lower bounded by $r_0$ which expresses the sharpness prior of the pooling function. A large $r_0$ encourages the learned saliency map to have less diffused modes. Section \ref{sec_exp} studies the effect of such a prior in both classification and weakly-supervised localization settings.
\\
\noindent \textbf{Lower-bounding the adaptation of $r$.} The adaptive nature of $\var{\mbox{LSE}-{\mbox{LBA}}}$ is inspired by the design of adaptive \mbox{GM} pooling proposed in \cite{adaptpool}. In addition to being numerically more stable in computation, our pooling function reparameterizes $r$ in $\mbox{LSE}$ pooling with $r = r_0 + \exp(\beta)$ where $r_0$ is a positive constant and $\beta$ a learnable parameter. It is easy to see that $r$ is lower bounded by $r_0$ which expresses the sharpness prior of the pooling function. A large $r_0$ encourages the learned saliency map to have less diffuse modes. Section \ref{sec_exp} studies the effect of such a prior in both classification and weakly-supervised localization settings.

\subsection{Weakly Supervised MIL Cost Function}
The overall information propagation path resembles that of a U-net from \cite{u_net}. The major architectural difference is that the proposed model extends U-nets to the weakly-supervised setting where pixel-wise labels are not available and only image-level annotations are utilized. 

Given the multi-resolution fused feature map at the highest level resolution $F^0(\mathbf{x}) \in \mathcal{R}^{w \times h \times c}$, it is further divided into a grid of $N \times N$ with N being the chosen resolution of the final saliency map. A typical choice is $N=w=h$, resulting in $F^0(\mathbf{x}) \in \mathcal{R}^{N \times N \times c}$. Each of the $N^2$ $c$-dimensional vectors represents an instance $I_n(\mathbf{x})$ in the bag $F^0$, where $n=\{1, \dots, N^2\}$. The $K$-class instance probability is $P(I_n(\mathbf{x}))=\sigmoid(W I_n(\mathbf{x}))$, where $W$ is a $K$ by $c$ parameter matrix that is shared among all $N^2$ instances. This leads to the final probabilistic saliency map $S(\mathbf{x}) \in \mathcal{Q}^{N \times N \times K}$ for $K$ classes. Following the pooling function introduced in Section \ref{sec_lselba}, we have $P(\mathbf{x})=\var{\mbox{LSE}-{\mbox{LBA}}}(S(\mathbf{x}))$. The global prediction $P(\mathbf{x})$ is a $K$-dimensional vector and represents, according to the probability-preserving property of $\var{\mbox{LSE}-{\mbox{LBA}}}$ pooling described in Section \ref{sec_lselba}, the probability of $\mathbf{x}$ belonging to $K$ classes. Hence, a standard multi-class cross-entropy cost can be directly computed given $\mathbf{y}$.

\section{Experiments}\label{sec_exp}

\subsection{Dataset}
In order to verify the efficacy of the proposed model in the context of medical diagnosis, experiments are conducted on the largest Chest X-ray dataset that is publicly available. The NIH Chest X-ray dataset was originally introduced in \cite{chestx8}. It contains 112,120 frontal-view chest X-rays taken from 30,805 patients, where 51,708 images contain at least one of 14 labeled pathologies. Although the original DICOM files and the accompanied radiologists' reports are not released, the images are made available in PNG format with a standardized spatial resolution of $1024 \times 1024$. Other clinical information including patients' age and gender are accessible in addition to the pathology labels. Such information, however, is not utilized for the sake of this study.

Following the original work of \cite{chestx8}, where this dataset was introduced, there are several notable studies that utilized the same dataset (\cite{LiLearning,chexnet,tienet,cascade,agcnn,feifei,2018arXiv180302315B}). Unfortunately, they each performed their own splits (except the very recent work of \cite{SebastianDenseNet} with performance reported using significantly more additional data) making it hard to establish consistent benchmarks. For our studies, we have adopted the official train and test splits \footnote{\url{https://nihcc.app.box.com/v/ChestXray-NIHCC}} from \cite{chestx8}. They are the only ones publicly available and we have confirmed that the splitting was done by patient rather than by image so that images of the same patient do not occur in both train and test at the same time. Table \ref{tab_data} contains some key statistics. The validation set is created by a further 75\%-25\% split of the official training set. In addition, manually annotated bounding boxes are provided for a small subset of images from the official test set.
\begin{table}[ht]
\caption{Number of training examples and number of bounding boxes by abnormality by split used in this work. Note that there are no bounding boxes available for training (75\% of the official training set) and validation sets (25\% of the official training set).}
\label{tab_data}
\begin{center}
\begin{tabular}{|c|c|c|c|c|c|c|c|c|c|c|}
\hline
\textbf{abnormality} & \textbf{train} & \textbf{valid} &\textbf{test}&\textbf{bbox}&\textbf{abnormality}&\textbf{train}&\textbf{valid}&\textbf{test}&\textbf{bbox}\\
\hline \hline
Atelectasis & 6,168 & 2,112 & 3,255 & 180&Cardiomegaly &  1,273& 434 & 1,065&146\\
Effusion & 6,537 & 2,122 & 4,648&153&Infiltration & 10,244 &3,538 & 6,088&123\\
Mass & 3,012 & 1,022 & 1,712&85&Nodule & 3,501 & 1,207&615&79\\
Pneumonia & 655 & 221& 477& 120&Pneumothorax & 1,939 & 698&2,661&98\\
Consolidation & 2,114 & 738&1,815 &-&Edema & 1,027 & 361& 925&-\\
Emphysema & 1,075 & 348& 1,093&- &Fibrosis & 963 & 288& 435&-\\
Pleural thickening & 1,693 & 549& 1,143&-& Hernia & 105 & 36&86&-\\
\hline
\end{tabular}
\end{center}
\end{table}
\subsection{Training Procedures and Evaluation Metrics}
For computational efficiency, the inputs of $1024 \times 1024$ are downsampled \footnote{with `skimage.transform.resize'} to $512 \times 512$. The loss of resolution may lead to sub-optimal results, but this is not the focus of this study. Data augmentation is applied during training in which each image is zoomed by a factor uniformly sampled from $\left[0.25, 0.75\right]$, translated in four directions by a factor uniformly sampled from $\left[-50, 50\right]$ pixels, and rotated by a factor uniformly sampled from $\left[-25, 25\right]$ degrees. After data augmentation, the inputs are normalized to the interval $\left[0, 1\right]$. To further regularize the model, a weight decay is applied with a coefficient of $10^{-5}$.
% Data augmentations are applied on each image during training, with a zooming factor uniformly sampled from 0.75 to 0.25, a translation factor in four directions uniformly from -50 to 50 pixels, a rotation factor uniformly from -25 to 25 degrees.

Models are trained from scratch using only the NIH training set with the Adam optimizer (\cite{kingma2014adam}) and a learning rate of 0.001. Early stopping is performed on the validation set based on the average AUC (Area Under the ROC curve) over all 14 pathologies. For classification, we report the AUC per abnormality on the test set with the best model chosen on the validation set.
\begin{table}[ht]
\caption{$\mbox{IoBB}$ is very sensitive to the choice of binarization threshold $\tau$ to discretize probabilistic saliency maps into binary foreground and background masks, demonstrated on three choices of $r_0$ on three abnormalities.}
\label{tab_iobb}
\begin{center}
\begin{tabular}{|c|c|c|c|c|c|c|}
\hline
\multicolumn{7}{|c|}{T(IoBB)=0.5} \\
\hline
&\multicolumn{2}{|c|}{\textbf{Pneumonia}} & \multicolumn{2}{|c|}{\textbf{Nodule}} & \multicolumn{2}{|c|}{\textbf{Infiltration}}\\
\hline
&$\tau=0.1$&$\tau=0.4$&$\tau=0.4$&$\tau=0.8$&$\tau=0.1$&$\tau=0.8$\\
\hline
\hline
$r_0=0$ &0.3334&0.3167&0.0506&0.1645&0.1382&0.4634\\
$r_0=5$ &0.5083&0.3500&0.1012&0.2025&0.1626&0.5284\\
$r_0=10$&0.4416&0.0500&0.1772&0.2152&0.1220&0.3008\\
\hline
\cite{chestx8}&\multicolumn{2}{|c|}{0.3833}&\multicolumn{2}{|c|}{0.0126}&\multicolumn{2}{|c|}{0.4227}\\
\hline
\end{tabular}
\end{center}
\end{table}

Similar to \cite{chestx8}, no bounding boxes are used at training time so that the model remains weakly supervised with respect to the task of localization. The best models on the classification task are then evaluated on their localization performance. Both \cite{feifei} and \cite{chestx8} report the quality of localization using the metric of intersection over detected bounding boxes ($\mbox{IoBB}$) with $\mbox{T}(\mbox{IoBB})=\alpha$, where $\alpha$ is set at a certain threshold. We have noticed that $\mbox{IoBB}$ is extremely sensitive to the choice of the discretization threshold by which the predicted probability score $S$ is binarized before being compared with ground truth bounding boxes. This is illustrated in Table \ref{tab_iobb} where the accuracies (defined in \cite{chestx8}) with $T(\mbox{IoBB})=0.5$ are computed with different binarization threshold $\tau$. We have deliberately chosen three types abnormalities that have distinct visual characteristics: nodule (mostly focalized), infiltration (mostly diffuse) and pneumonia (either focalized or diffuse). Although all three of them have achieved significantly better performance compared with \cite{chestx8} ($\tau$ unspecified), the performance is highly sensitive to the choice of $\tau$, making this metric unsuitable for benchmarking, especially in the weakly supervised learning setting where the entropy of the resulting probabilistic saliency maps are typically high.

We therefore resort to the continuous version of $\mbox{DICE}=(2 \times \mbox{S}\times \mbox{G})/(\mbox{S}^2 + \mbox{G}^2)$ where $S$ is the probabilistic saliency map directly output by the model, and $G$ the ground truth binary bounding box downsampled to $512 \times 512$, the same resolution as the model input. Such a metric has been widely adopted as the standard cost function for training semantic segmentation models (\cite{vnet}). It naturally takes into account the probability while avoiding the decision of having to select the discretization threshold $\tau$, often an arbitrary decision without a separate validation set with bounding boxes.
\subsection{Quantitative Results}
\begin{table}[ht]
\caption{Abnormality classification and localization performance on 14 abnormalities on NIH Chest X-ray test set. Three models with different lower-bounded adaptation $r_0$ (Section \ref{sec_mil}) are included. The impact of $r_0$ is much more pronounced in localization than in classification. The model is only trained on NIH data while \cite{chestx8} uses a pre-trained model on ImageNet without multi-resolution fusion. Bolded numbers indicate the maxima other than statistical significance.}
\label{tab_auc_dice}
\begin{center}
\begin{tabular}{|c||c||c|c|c|c|c|c|}
\hline
\multirow{2}{*}{} & \multicolumn{4}{|c|}{AUC} &\multicolumn{3}{|c|}{DICE}\\
\cline{2-8}
&\cite{chestx8}&$r_0=0$&$r_0=5$&$r_0=10$&$r_0=0$&$r_0=5$&$r_0=10$\\
\cline{1-8}
\textbf{Atelectasis}       & 0.7003&\textbf{0.733}&0.728&0.724&0.204&\textbf{0.240}&0.211\\
\textbf{Cardiomegaly}      & 0.8100&0.856&\textbf{0.858}&0.854&\textbf{0.180}&0.114&0.076\\
\textbf{Effusion}          & 0.7585&\textbf{0.806}&0.803&0.795&0.293&\textbf{0.294}&0.242\\
\textbf{Infiltration}      & 0.6614&0.673&\textbf{0.675}&0.668&\textbf{0.325}&0.312&0.286\\
\textbf{Nodule}            & 0.6687&0.718&0.724&\textbf{0.727}&0.202&\textbf{0.238}&0.196\\
\textbf{Mass}              & 0.6933&0.777&0.777&\textbf{0.778}&\textbf{0.295}&\textbf{0.295}&0.241\\
\textbf{Pneumonia}         & 0.6580&0.684&\textbf{0.690}&0.687&\textbf{0.112}&0.104&0.072\\
\textbf{Pneumothorax}      & 0.7993&\textbf{0.805}&0.791&0.763&\textbf{0.039}&0.023&0.028\\
\textbf{Consolidation}     & 0.7032&0.711&0.714&\textbf{0.717}&-&-&-\\
\textbf{Edema}             & 0.8052&\textbf{0.806}&0.804&0.801&-&-&-\\
\textbf{Emphysema}         & 0.8330&\textbf{0.842}&0.822&0.771&-&-&-\\
\textbf{Fibrosis}          & \textbf{0.7859}&0.743&0.757&0.731&-&-&-\\
\textbf{Pleural thickening}& 0.6835&\textbf{0.724}&0.715&0.712&-&-&-\\
\textbf{Hernia}            & \textbf{0.8717}&0.775&0.764&0.824&-&-&-\\
\hline
A.V.G.                     & 0.738 &0.761&0.760&0.754&-&-&-\\
\hline
\end{tabular}
\end{center}
\end{table}
Table \ref{tab_auc_dice} summarizes both the classification and weakly supervised localization performance of the proposed model. With different choice of $r_0$ it outperforms the current state-of-the-art in 9 of 14 abnormalities by significant margins. It is on-par with the current state-of-the-art on 3 of the rest 5 abnormalities, although being able to do so without using any extra data. On fibrosis and hernia, it is not able to match the existing model pre-trained on ImageNet, indicating that for those two abnormalities, pre-training would be able to bring a much more significant benefit, especially considering the fact that there are only about 1,000 training example for those two classes combined (Table \ref{tab_data}). One significant trend is that the final AUCs are robust to the choice of $r_0$, except in emphysema, fibrosis and hernia with the optimal choice of $r_0=0$, $r_0=5$ and $r_0=10$ respectively.

Compared with classification, the choice of $r_0$ has a more significant impact on abnormality localization due to their likely distinct visual appearance. For instance, when $r_0$ is small and the sharpness prior is weak, a model tends to perform well on visually diffused abnormalities such as cardiomegaly, infiltration and pneumonia. As one strengthens the sharpness prior, localization of focalized and patchy abnormalities is improved, as in the case of atelectasis and nodule. When choosing $r_0$ to be too large, the performance of diffused abnormalities degrades, such as atelectasis, cardiomegaly, effusion and pneumonia. 
\subsection{Qualitative Results}

\begin{figure}[ht]
\centering
\includegraphics[scale=0.17]{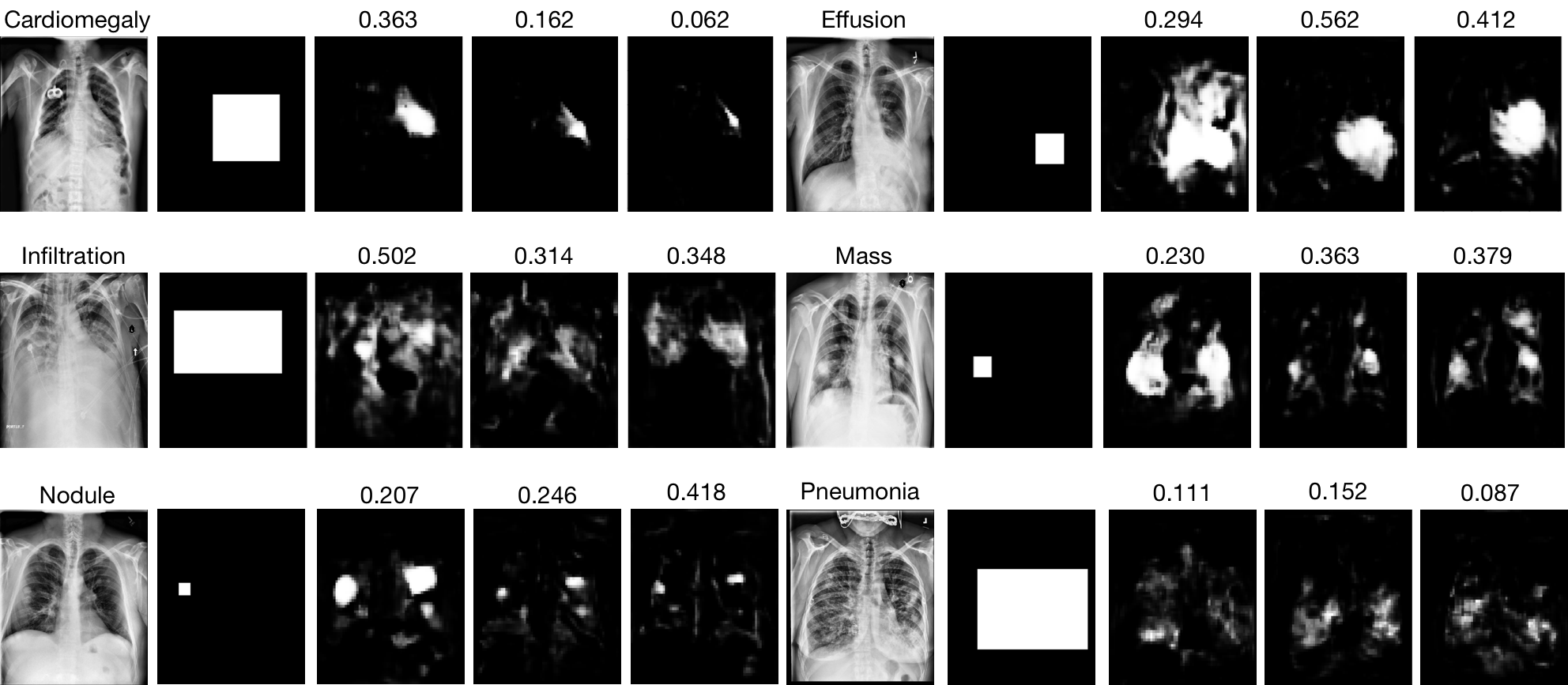}
\caption{Some examples of generated saliency maps with their corresponding DICE score at the top. Columns from the left to the right: image, ground truth bounding box, $r_0=0$, $r_0=5$, $r_0=10$.}
\label{fig_heatmaps}
\end{figure}

Figure \ref{fig_heatmaps} contains the model-generated saliency maps along with the original images and ground truth bounding boxes. To accompany the qualitative evaluation we also include their corresponding DICE score computed with respect to the ground truth shown. Visualization confirms that increasing $r_0$ results in overall sharper saliency maps. It can be argued that using bounding boxes to delineate abnormalities is limited by over-estimating their true ROIs, which is evident in the cases of infiltration and pneumonia. It can also be observed from the figure that some model findings are incorrectly marked as false positives due to labeling noise wherein the ground-truth reader missed the finding.

\section{Conclusions}
Modern medical imaging has imposed upon machine learning several fundamental challenges in disease diagnosis and localization. This work has studied a partial solution through the novel combination of multi-resolution and multi-instance learning with a customized pooling function designed to deliver more accurate diagnosis and higher-resolution pathology-based saliency maps. The proposal is evaluated on the largest publicly available chest X-ray dataset. We have shown both quantitatively and qualitatively its efficacy, establishing the new state-of-the-art performance on a majority of the 14 abnormalities. Experimental results further suggest a future study where the proposed model may be pre-trained on data from another domain to improve performance.

\bibliography{eccv2018submission}

\begin{thebibliography}{10}

\bibitem{dietterich1997solving}
Dietterich, T.G., Lathrop, R.H., Lozano-P{\'e}rez, T.:
\newblock Solving the multiple instance problem with axis-parallel rectangles.
\newblock Artificial intelligence \textbf{89}(1-2) (1997)  31--71

\bibitem{amores2013multiple}
Amores, J.:
\newblock Multiple instance classification: Review, taxonomy and comparative
  study.
\newblock Artificial Intelligence \textbf{201} (2013)  81--105

\bibitem{chestx8}
Wang, X., Peng, Y., Lu, L., Lu, Z., Bagheri, M., Summers, R.M.:
\newblock Chestx-ray8: Hospital-scale chest x-ray database and benchmarks on
  weakly-supervised classification and localization of common thorax diseases.
\newblock In: 2017 {IEEE} Conference on Computer Vision and Pattern
  Recognition, {CVPR} 2017, Honolulu, HI, USA, July 21-26, 2017. (2017)
  3462--3471

\bibitem{feifei}
Li, Z., Wang, C., Han, M., Xue, Y., Wei, W., Li, L., Li, F.:
\newblock Thoracic disease identification and localization with limited
  supervision.
\newblock CoRR \textbf{abs/1711.06373} (2017)

\bibitem{courtiol2018classification}
Courtiol, P., Tramel, E.W., Sanselme, M., Wainrib, G.:
\newblock Classification and disease localization in histopathology using only
  global labels: A weakly-supervised approach.
\newblock arXiv preprint arXiv:1802.02212 (2018)

\bibitem{u_net}
Ronneberger, O., Fischer, P., Brox, T.:
\newblock U-net: Convolutional networks for biomedical image segmentation.
\newblock CoRR \textbf{abs/1505.04597} (2015)

\bibitem{pyramids}
Adelson, E.H., Anderson, C.H., Bergen, J.R., Burt, P.J., Ogden, J.M.:
\newblock {1984, Pyramid methods in image processing}.
\newblock RCA Engineer \textbf{29}(6) (1984)  33--41

\bibitem{hypercolumns}
Hariharan, B., Arbel{\'{a}}ez, P.A., Girshick, R.B., Malik, J.:
\newblock Hypercolumns for object segmentation and fine-grained localization.
\newblock CoRR \textbf{abs/1411.5752} (2014)

\bibitem{fcn}
Long, J., Shelhamer, E., Darrell, T.:
\newblock Fully convolutional networks for semantic segmentation.
\newblock CoRR \textbf{abs/1411.4038} (2014)

\bibitem{class_activation_maps}
Zhou, B., Khosla, A., Lapedriza, {\`{A}}., Oliva, A., Torralba, A.:
\newblock Learning deep features for discriminative localization.
\newblock CoRR \textbf{abs/1512.04150} (2015)

\bibitem{refine_net}
Lin, G., Milan, A., Shen, C., Reid, I.D.:
\newblock Refinenet: Multi-path refinement networks for high-resolution
  semantic segmentation.
\newblock CoRR \textbf{abs/1611.06612} (2016)

\bibitem{seg_net}
Badrinarayanan, V., Kendall, A., Cipolla, R.:
\newblock Segnet: {A} deep convolutional encoder-decoder architecture for image
  segmentation.
\newblock CoRR \textbf{abs/1511.00561} (2015)

\bibitem{res_net}
He, K., Zhang, X., Ren, S., Sun, J.:
\newblock Deep residual learning for image recognition.
\newblock CoRR \textbf{abs/1512.03385} (2015)

\bibitem{tiramisu}
J{\'{e}}gou, S., Drozdzal, M., V{\'{a}}zquez, D., Romero, A., Bengio, Y.:
\newblock The one hundred layers tiramisu: Fully convolutional densenets for
  semantic segmentation.
\newblock CoRR \textbf{abs/1611.09326} (2016)

\bibitem{weaklearning}
Zhou, Z.H.:
\newblock A brief introduction to weakly supervised learning.
\newblock NSR \textbf{5.1} (2017)  44--53

\bibitem{detectors_emerge}
Zhou, B., Khosla, A., Lapedriza, {\`{A}}., Oliva, A., Torralba, A.:
\newblock Object detectors emerge in deep scene cnns.
\newblock CoRR \textbf{abs/1412.6856} (2014)

\bibitem{fcnmil}
Pathak, D., Shelhamer, E., Long, J., Darrell, T.:
\newblock Fully convolutional multi-class multiple instance learning.
\newblock CoRR \textbf{abs/1412.7144} (2014)

\bibitem{im2pix}
Pinheiro, P.H.O., Collobert, R.:
\newblock From image-level to pixel-level labeling with convolutional networks.
\newblock In: {IEEE} Conference on Computer Vision and Pattern Recognition,
  {CVPR} 2015, Boston, MA, USA, June 7-12, 2015. (2015)  1713--1721

\bibitem{overfeat}
Sermanet, P., Eigen, D., Zhang, X., Mathieu, M., Fergus, R., LeCun, Y.:
\newblock Overfeat: Integrated recognition, localization and detection using
  convolutional networks.
\newblock CoRR \textbf{abs/1312.6229} (2013)

\bibitem{babenko}
Babenko, B.:
\newblock Multiple instance learning: Algorithms and applications.
\newblock (2009)

\bibitem{microscopy}
Kraus, O.Z., Ba, L.J., Frey, B.J.:
\newblock Classifying and segmenting microscopy images with deep multiple
  instance learning.
\newblock Bioinformatics \textbf{32}(12) (2016)  52--59

\bibitem{mammomil}
Zhu, W., Lou, Q., Vang, Y.S., Xie, X.:
\newblock Deep multi-instance networks with sparse label assignment for whole
  mammogram classification.
\newblock In: Medical Image Computing and Computer Assisted Intervention -
  {MICCAI} 2017 - 20th International Conference, Quebec City, QC, Canada,
  September 11-13, 2017, Proceedings, Part {III}. (2017)  603--611

\bibitem{attentionmil}
Ilse, M., Tomczak, J.M., Welling, M.:
\newblock Attention-based deep multiple instance learning.
\newblock CoRR \textbf{abs/1802.04712} (2018)

\bibitem{gradcam}
Selvaraju, R.R., Cogswell, M., Das, A., Vedantam, R., Parikh, D., Batra, D.:
\newblock Grad-cam: Visual explanations from deep networks via gradient-based
  localization.
\newblock In: {IEEE} International Conference on Computer Vision, {ICCV} 2017,
  Venice, Italy, October 22-29, 2017. (2017)  618--626

\bibitem{seedexpand}
Kolesnikov, A., Lampert, C.H.:
\newblock Seed, expand and constrain: Three principles for weakly-supervised
  image segmentation.
\newblock In: Computer Vision - {ECCV} 2016 - 14th European Conference,
  Amsterdam, The Netherlands, October 11-14, 2016, Proceedings, Part {IV}.
  (2016)  695--711

\bibitem{vnet}
Milletari, F., Navab, N., Ahmadi, S.:
\newblock V-net: Fully convolutional neural networks for volumetric medical
  image segmentation.
\newblock In: Fourth International Conference on 3D Vision, 3DV 2016, Stanford,
  CA, USA, October 25-28, 2016. (2016)  565--571

\bibitem{deep_weak_supervision}
Jia, Z., Huang, X., Chang, E.I., Xu, Y.:
\newblock Constrained deep weak supervision for histopathology image
  segmentation.
\newblock {IEEE} Trans. Med. Imaging \textbf{36}(11) (2017)  2376--2388

\bibitem{chexnet}
Rajpurkar, P., Irvin, J., Zhu, K., Yang, B., Mehta, H., Duan, T., Ding, D.,
  Bagul, A., Langlotz, C., Shpanskaya, K., Lungren, M.P., Ng, A.Y.:
\newblock Chexnet: Radiologist-level pneumonia detection on chest x-rays with
  deep learning.
\newblock CoRR \textbf{abs/1711.05225} (2017)

\bibitem{LiLearning}
Yao, L., Poblenz, E., Dagunts, D., Covington, B., Bernard, D., Lyman, K.:
\newblock Learning to diagnose from scratch by exploiting dependencies among
  labels.
\newblock CoRR \textbf{abs/1710.10501} (2017)

\bibitem{agcnn}
Guan, Q., Huang, Y., Zhong, Z., Zheng, Z., Zheng, L., Yang, Y.:
\newblock Diagnose like a radiologist: Attention guided convolutional neural
  network for thorax disease classification.
\newblock CoRR \textbf{abs/1801.09927} (2018)

\bibitem{cascade}
Kumar, P., Grewal, M., Srivastava, M.M.:
\newblock Boosted cascaded convnets for multilabel classification of thoracic
  diseases in chest radiographs.
\newblock CoRR \textbf{abs/1711.08760} (2017)

\bibitem{tienet}
Wang, X., Peng, Y., Lu, L., Lu, Z., Summers, R.M.:
\newblock Tienet: Text-image embedding network for common thorax disease
  classification and reporting in chest x-rays.
\newblock CoRR \textbf{abs/1801.04334} (2018)

\bibitem{wu2016wider}
Wu, Z., Shen, C., Hengel, A.v.d.:
\newblock Wider or deeper: Revisiting the resnet model for visual recognition.
\newblock arXiv preprint arXiv:1611.10080 (2016)

\bibitem{huang2017densely}
Huang, G., Liu, Z., Weinberger, K.Q., van~der Maaten, L.:
\newblock Densely connected convolutional networks.
\newblock In: Proceedings of the IEEE conference on computer vision and pattern
  recognition. Volume~1. (2017) ~3

\bibitem{adaptpool}
Liu, D., Zhou, Y., Sun, X., Zha, Z., Zeng, W.:
\newblock Adaptive pooling in multi-instance learning for web video annotation.
\newblock In: 2017 {IEEE} International Conference on Computer Vision
  Workshops, {ICCV} Workshops 2017, Venice, Italy, October 22-29, 2017. (2017)
  318--327

\bibitem{2018arXiv180302315B}
{Baltruschat}, I.M., {Nickisch}, H., {Grass}, M., {Knopp}, T., {Saalbach}, A.:
\newblock {Comparison of Deep Learning Approaches for Multi-Label Chest X-Ray
  Classification}.
\newblock ArXiv e-prints (March 2018)

\bibitem{SebastianDenseNet}
Guendel, S., Grbic, S., Georgescu, B., Zhou, K., Ritschl, L., Meier, A.,
  Comaniciu, D.:
\newblock Learning to recognize abnormalities in chest x-rays with
  location-aware dense networks.
\newblock CoRR (2018)

\bibitem{kingma2014adam}
Kingma, D.P., Ba, J.:
\newblock Adam: A method for stochastic optimization.
\newblock arXiv preprint arXiv:1412.6980 (2014)

\end{thebibliography}
\bibliographystyle{splncs}

\end{document}